%% file: main.tex
\documentclass[10pt,twocolumn,letterpaper]{article}

\usepackage[pagenumbers]{cvpr} 

\usepackage{multirow}
\usepackage[accsupp]{axessibility}  

\definecolor{cvprblue}{rgb}{0.21,0.49,0.74}
\usepackage[pagebackref,breaklinks,colorlinks,allcolors=cvprblue]{hyperref}

\def\paperID{34} 
\def\confName{CVPR}
\def\confYear{2026}

\title{ReLeaf: Benchmarking Leaf Segmentation across Domains and Species}

\author{Robert Martinko$^{1,2}$ ~~~ Daniel Steininger$^{1}$ ~~~ Julia Simon$^{1}$ ~~~ Andreas Trondl$^{1}$ ~~~ Matthias Blaickner$^{2}$ \\
$^{1}$AIT Austrian Institute of Technology, Center for Vision, Automation \& Control \\
$^{2}$University of Applied Sciences Technikum Wien, Computer Science \& Applied Mathematics \\
{\tt\small\{robert.martinko,daniel.steininger,julia.simon,andreas.trondl.fl\}@ait.ac.at} \\
\tt\small{matthias.blaickner@technikum-wien.at}}

\begin{document}

\maketitle
\input{sec/0_abstract}    
\input{sec/1_introduction}
\input{sec/2_related_work}
\input{sec/3_datasets}
\input{sec/4_methodology}
\input{sec/5_results}
\input{sec/6_conclusion}

{
    \small
    \bibliographystyle{ieeenat_fullname}
    \bibliography{main}
}

\input{sec/X_suppl}

\end{document}

%% file: sec/0_abstract.tex
\begin{abstract}
Rising global food demand and growing climate pressure increase the need for sustainable, precise agricultural practices. Automated, individualized plant treatment relies on fine‑grained visual analysis, yet leaf‑level segmentation remains underexplored despite its value for assessing crop health, growth dynamics, yield potential and localized stress symptoms. Progress is limited by a lack of dedicated datasets, especially regarding species coverage, and by the absence of systematic evaluations of modern instance‑segmentation architectures for this task. We address these gaps by surveying current data and identifying four suitable, publicly available leaf‑segmentation datasets. Using them, we compare one‑stage, two‑stage and Transformer-based detectors and identify a YOLO26 model configuration to provide the best trade‑off for real‑world precision‑agriculture tasks. Extensive cross‑domain generalization experiments reveal substantial performance drops across plant species and recording setups, especially for models trained solely on laboratory data. To strengthen data availability, we introduce a new benchmark dataset with leaf‑level masks for 23 plant species, created via semi‑automatic annotation of selected CropAndWeed images. A model trained on all four existing datasets achieves a mean mAP\textsuperscript{50-95} of 83.9\% across their corresponding test sets and 40.2\% on our new benchmark, demonstrating improved generalization and highlighting the need for diverse leaf-segmentation datasets in robust precision agriculture.
\end{abstract}

%% file: sec/1_introduction.tex
\section{Introduction}

\begin{figure}
    \centering
    \includegraphics[width=1.0\columnwidth]{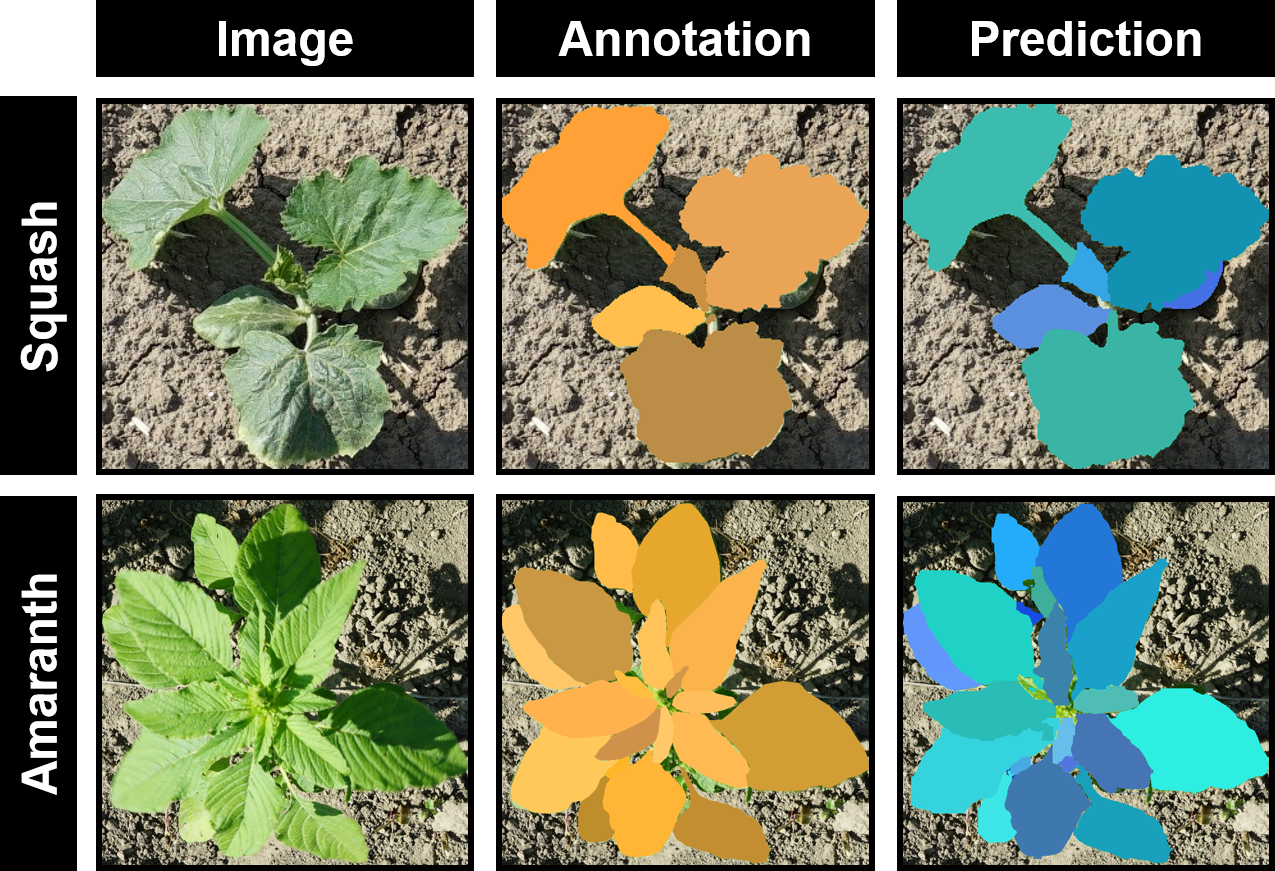}
    \caption{Representative samples from \textit{CropAndWeed} \cite{StTr23} and leaf-level annotations for our \textit{CropAndWeedAndLeaf} benchmark, along with results of the multi-domain \textit{ReLeaf} model, illustrating its generalization to unseen species.}
    \label{fig:overview_results}
\end{figure}

As global food demand continues to rise and climate conditions intensify, agriculture faces the dual challenge of increasing productivity while reducing resource consumption. Precision agriculture offers a promising path by enabling targeted, plant‑level interventions that minimize fertilizer, pesticide and water requirements. Such individualized treatment relies on automated image‑based analysis, yet variable field conditions across seasons and sensor setups make robust performance difficult and demand models with strong cross‑domain generalization. While plant‑level classification and detection have progressed notably, pixel‑level segmentation provides more fine‑grained morphological information most useful for plant‑specific decision‑making \cite{CaOz19, StTr23}.

Within this context, leaf segmentation is both particularly complex and comparatively underexplored. Segmenting individual leaves reveals morphological traits that indicate plant health \cite{DeHa17}, growth stage and yield potential \cite{GiVa16}, and enables computation of the Leaf Area Index \cite{FaBa19, ItKe18}. Leaf‑level masks also facilitate early detection of localized stress symptoms such as fungal infections or drought damage \cite{DeHa17, SiVi17, SaMe18}. Beyond field crops, robust methods for fine‑grained part segmentation are relevant for related domains such as autonomous fruit‑harvesting systems \cite{LeSa16} or industrial inspection of overlapping objects.

\begin{figure*}
    \centering
    \includegraphics[width=2.0\columnwidth]{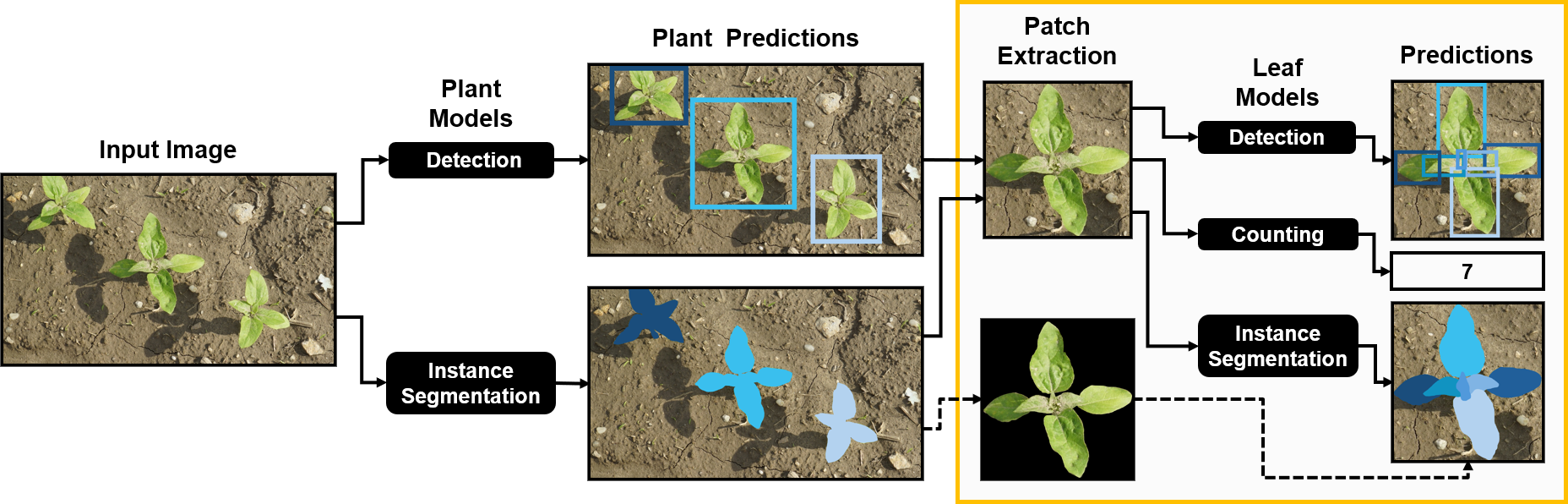}
    \caption{Representative processing pipeline, illustrating the typical context of leaf segmentation. Image patches depicting individual detected plants are cropped from larger-resolution field recordings for fine-grained analysis. While the upstream tasks of plant-level detection and instance segmentation are well researched for common species, tasks related to leaf-segmentation remain underrepresented.}
    \label{fig:processing_pipeline}
\end{figure*}

Although modern model architectures perform strongly on large‑scale benchmarks like MS COCO \cite{LiMa14} and ImageNet \cite{DeDo09}, their accuracy often degrades on agricultural imagery due to substantial domain shift. Only limited work compares state‑of‑the‑art architectures specifically for leaf‑level segmentation, which is an important gap since real‑time robotics benefit from efficient one‑stage detectors such as YOLO26 \cite{JoCh24}, while two‑stage systems like Detectron2 \cite{WuKi19} can better capture fine‑grained, overlapping structures. Transformer‑based models, such as RF‑DETR \cite{RoRo25}, may handle long‑range dependencies but remain challenging for real‑time deployment. Progress is further limited by the scarcity of dedicated datasets due to the difficulty and cost of producing pixel‑accurate leaf masks (\cref{fig:overview_results}).

To address these gaps, this work evaluates the generalizability of modern deep‑learning architectures and datasets for leaf segmentation and expands data availability for this task by proposing the following key contributions:

\begin{itemize}
    \setlength\itemsep{1em}
    \item 
        We evaluate the applicability of three representative deep-learning architectures for leaf-segmentation tasks, comparing segmentation performance and runtime efficiency across multiple configurations.   
    \item 
        We conduct extensive cross-domain generalization experiments to quantify the adaptation potential of leaf-segmentation models trained on four selected datasets, covering real-world and laboratory settings.  
    \item 
        We introduce a novel benchmark dataset created by semi-automatically annotating a selected subset of the \textit{CropAndWeed} \cite{StTr23} dataset with leaf-segmentation masks for 23 plant species, allowing a comprehensive analysis of model generalization.\footnote{Dataset, code and models are available for academic use at \href{https://github.com/cropandweed/releaf}{https://github.com/cropandweed/releaf}}
\end{itemize}

%% file: sec/2_related_work.tex
\section{Related work}
Automated phenotyping has undergone major changes as traditional feature‑ or threshold‑based methods reached their limits \cite{AiSh17}, particularly due to the growing need for precise leaf‑level analysis \cite{MeZe17, WeMa22}. Deep‑learning approaches have since become the standard \cite{RoGi22}, with convolutional neural networks demonstrating strong performance in learning complex plant patterns directly from raw imagery \cite{LiRu21}.

Early deep‑learning phenotyping relied heavily on semantic segmentation, with U‑Net \cite{RoOl15} emerging as the dominant architecture. Numerous studies \cite{FaMu19, WeMa22, StTr23} showed that U‑Net variants reliably separate leaf regions from background, even in challenging soil conditions. However, semantic approaches struggle when leaf‑level traits are required, as overlapping leaves are merged into a single region \cite{KuZv19, RoGi22}. To address this limitation, instance‑segmentation models have gained traction. Mask R‑CNN, for example, has been shown to better separate overlapping leaves \cite{KuZv19, WeMa22}, albeit with higher computational demands \cite{LiRu21}.

The adoption of instance segmentation accelerated with modular frameworks like Detectron2 \cite{WuKi19}, which provides high‑performance Mask R‑CNN implementations widely used as baselines in agricultural research. To reduce computational cost, other studies \cite{WaDe24, BoZh19, HuMo22, WaMi25} explored one‑stage detectors such as the YOLO family \cite{JoCh24}, which offer substantial speed advantages enabling real‑time field‑robot applications. However, CNN-based detectors remain limited by their local receptive fields. Recent computer‑vision trends increasingly rely on Vision Transformers for global context modeling \cite{ZhSi21, CaNi20, ZhYi24} or foundation models \cite{WaZh25}, which are, however, hardly compatible with the demands of real-time field applications. Although state‑of‑the‑art models demonstrate impressive performance on general-purpose data, results degrade sharply when applied to precision‑agriculture tasks due to strong domain shifts \cite{BeEn20}.

To address these challenges, our work systematically evaluates how representative detectors perform on leaf‑level instance segmentation and examines their generalization across datasets and recording conditions.

%% file: sec/3_datasets.tex
\begin{table*}
    \caption{Key statistics of selected, excluded and newly annotated datasets: \textbf{Anno}tations of leaves (either per \textit{Plant}, per \textit{Image} or as classification labels for leaf \textit{Count}ing), numbers of \textbf{Plants} in the original dataset and \textbf{Selected} patches with a a minimum area of 600 pixels (unknown for \textit{Airphen} due to image-level leaf annotations), numbers of plant \textbf{Species}, median original \textbf{Image size}s, median cropped \textbf{Patch size}s of selected instances, recording \textbf{Sensors} and \textbf{Setting}s.}
    \label{tab:dataset_comparison}
    \centering
    \begin{tabular}{@{}lccccccc@{}}
        \toprule
        \textbf{Dataset} & \textbf{Anno} & \textbf{Plants (Selected)} & \textbf{Species} & \textbf{Image size} & \textbf{Patch size} & \textbf{Sensors} & \textbf{Setting} \\
        \midrule
        \textit{LSC} \cite{ScMi14} & Plant & 810 (707) & 2 & 441$\times$441 & 274$\times$241 & RGB & Lab \\
        \textit{Komatsuna} \cite{UcSa17} & Plant & 1,200 (1,055) & 1 & 480$\times$480 & 176$\times$142 & RGB-D & Lab \\
        \textit{GrowliFlower} \cite{KiJu23} & Plant & 2,587 (2,432) & 1 & 448$\times$368 & 154$\times$151 & RGB & Real \\
        \textit{PhenoBench} \cite{WeJa24} & Plant & 23,062 (13,752) & 1 & 1,024$\times$1,024 & 152$\times$141 & RGB & Real \\
        \midrule
        \textit{WeedGrowthState} \cite{TeNi18} & Count & 9,372 & 18 & - & 205$\times$205 & RGB & Real \\
        \textit{Aberystwyth} \cite{BeJo16} & Image & 916 & 1 & 2,560$\times$1,920 & - & RGB & Lab \\
        \textit{Airphen} \cite{VaJe22} & Image & - & 1 & 1,196$\times$805 & - & RGB-IR & Real \\
        \midrule
        \textit{CropAndWeedAndLeaf} (Ours) & Plant & 8,819 (345) & 23 & 1,920$\times$1,088 & 286$\times$201 & RGB & Real \\
        \bottomrule
    \end{tabular}
\end{table*}

\section{Leaf-segmentation datasets}
In realistic applications, the task of leaf segmentation is usually embedded in a processing pipeline similar to \cref{fig:processing_pipeline}, where images of crops in the field are first processed by detection or instance-segmentation models to extract image patches corresponding to individual crops or weeds. These can then be analyzed on a fine-grained level producing downstream outputs, such as bounding boxes, masks or counts of individual leaves. While the tasks of plant-level detection and instance-segmentation have been addressed by numerous works, delivering robust results for many common plant species \cite{KhSh25, LeYa24}, learning tasks related to individual leaves remain strongly underrepresented.

Likewise, publicly available precision-agriculture datasets are increasingly abundant for tasks such as classification, detection and segmentation, including high variability regarding plant species, environmental conditions and sensor setups \cite{StTr23, WeMa22, CeKu25}. However, despite the growing importance of leaf-level analysis, data for this task remains drastically underrepresented. Accordingly, our extensive survey of available datasets providing leaf-segmentation annotations for field crops and corresponding weeds results in only seven potential sources. The following subsections provide details regarding their properties and selection, as well as a description of our own contribution to mitigating this data gap.

\subsection{Data analysis}
\begin{figure}
    \centering
    \includegraphics[width=1.0\columnwidth]{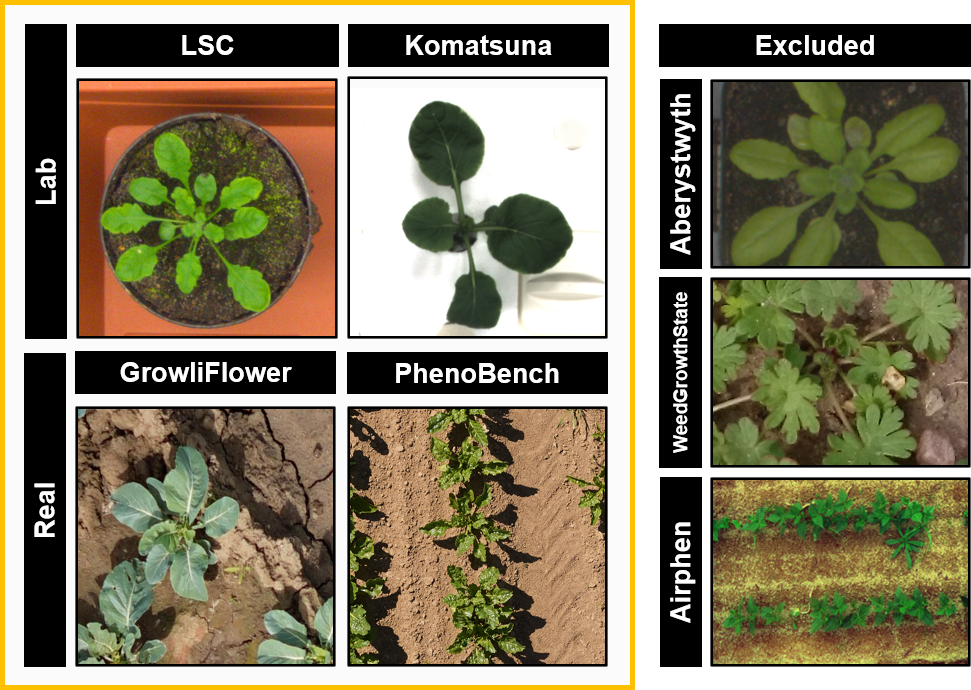}
    \caption{Representative images from leaf-segmentation datasets, depicting \textbf{Real}istic field settings \cite{KiJu23, WeJa24} and controlled \textbf{Lab}oratory environments \cite{ScMi14, UcSa17}, or \textbf{Excluded} due to incompatible leaf annotations \cite{BeJo16, TeNi18, VaJe22}.}
    \label{fig:related_datasets}
\end{figure}

We identified seven datasets potentially suitable for fine-grained leaf segmentation under field conditions, which are listed with representative samples in \cref{fig:related_datasets}. As visible in \cref{tab:dataset_comparison}, the quantity and quality of images and annotations vary significantly between these datasets. 

One important distinction is their recording setup. \textit{LSC} \cite{ScMi14}, \textit{Komatsuna} \cite{UcSa17} and \textit{Aberystwyth} \cite{BeJo16} use controlled laboratory settings, which facilitate cultivating the plants under constant environmental conditions and recording them with controlled lighting and specialized sensor setups, such as RGB-D cameras in the \textit{Komatsuna} dataset or high-frequency top-down recordings in \textit{Aberystwyth}. In contrast, \textit{PhenoBench} \cite{WeJa24}, \textit{GrowliFlower} \cite{KiJu23}, \textit{Airphen} \cite{VaJe22} and \textit{WeedGrowthState} \cite{TeNi18} were captured under natural field conditions, which present challenges such as weed pressure, varying soil conditions and fluctuating light levels. While \textit{PhenoBench} relies on UAV RGB imagery, \textit{GrowliFlower} and \textit{Airphen} integrate multispectral sensors (e.g., 6-band or RGB-IR) to improve the spectral separation between vegetation and background. This furthermore influences available recording resolutions. While the \textit{Aberystwyth} dataset operates at high image sizes of up to 2,560$\times$1,920 pixels, the field datasets have significantly lower resolutions. Additionally, the examined datasets vary widely regarding their numbers of images and plant instances. \textit{PhenoBench} provides 23,062 individual instances and thereby clearly surpasses all other datasets, which, however, still provide sufficient numbers to be useful for our proposed experiments.

The datasets furthermore differ in the annotations they include. Beyond leaf segmentation, \textit{PhenoBench}, \textit{GrowliFlower}, \textit{Komatsuna} and \textit{LSC} support the full range of learning tasks from classification and detection to plant-level instance segmentation, while \textit{Airphen} and \textit{Aberystwyth} are limited to semantic-segmentation tasks, yet provide additional annotation modalities. \textit{Aberystwyth} uses a color-coding system which indicates the growth phase according to the order of the leaves, thus facilitating the modeling of plant development. \textit{Airphen} offers a unique edge-based separation between inner and outer leaves to address challenges in dense vegetation. \textit{WeedGrowthState} takes a completely different approach, not focusing on geometry but on classical classification tasks, and labels plants in nine different growth stages based on their numbers of leaves. 

For the purpose of this work, only leaf-level instance-segmentation masks are relevant. However, they need to be assignable to individual plant instances, since we plan to evaluate patch-based approaches compatible with the pipeline described in \cref{fig:processing_pipeline}. \textit{LSC} and \textit{Komatsuna} explicitly provide cropped image patches for each plant with corresponding leaf annotations, while \textit{GrowliFlower} and \textit{PhenoBench} allow their automatic inference by providing plant-level instance masks for images containing multiple plants. \textit{Airphen} and \textit{Aberystwyth}, on the other hand, only provide leaf masks on the level of entire images without differentiating individual plants, while \textit{WeedGrowthState} is only annotated for leaf counting and does not provide any kind of masks.

\subsection{Data unification and aggregation}
Based on our dataset survey, we select \textit{PhenoBench}, \textit{GrowliFlower}, \textit{LSC} and \textit{Komatsuna} as the only leaf-segmentation datasets that meet the requirements for evaluating this task on image patches containing a single plant instance. To enable a consistent comparison across these heterogeneous data sources, we created a unified aggregation pipeline to normalize all inputs into a common format. All selected leaf-segmentation datasets are converted into image patches tightly cropped on individual plants to closely resemble the output of the detection stage in \cref{fig:processing_pipeline}. To ensure adequate spatial resolution for downstream analysis, only patches with areas of at least 600 pixels are included in the training and test data. This results in a total of 17,946 image patches across all four datasets, as visible in \cref{tab:dataset_comparison}.

\subsection{Benchmark dataset creation}
Our survey shows that leaf-level annotations are extremely scarce in existing public datasets and limited to only four plant species (sugar beet, cauliflower, tobacco and arabidopsis) in the datasets suitable for patch-based applications. To mitigate this gap and extend our quantitative evaluation to a wider set of species, we enriched a selected subset of the established \textit{CropAndWeed} \cite{StTr23} dataset with high-quality, leaf-level segmentation masks using a semi-automatic annotation approach, thereby creating a novel \textit{CropAndWeedAndLeaf} benchmark dataset. To select suitable data, we first extract image patches of all plant instances from \textit{CropAndWeed} with a minimum instance size of 128 pixels along both dimensions. To limit label noise and distribution shift, we exclude species with fewer than 15 retained instances and species exhibiting pronounced inter-class variability relative to the training distribution, yielding a curated set of 23 species (fully listed in the supplementary material) and a pool of 8,819 instances meeting our criteria. To balance statistical significance for benchmarking and manual annotation efforts, we randomly sample 15 image patches per species from the filtered data. This results in a total number of 345 samples in our dataset, evenly distributed across the included classes. For annotation, we developed a scalable pipeline that couples model-assisted pre-annotation with targeted expert refinement. A preliminary model from our ablation study is applied to generate initial leaf masks that are subsequently verified and corrected by human annotators in CVAT \cite{Cv26}. Typical edits include resolving artifacts caused by low resolution, handling pronounced phenotype variability, recovering tiny leaves missed by the model and splitting adjacent leaves that were incorrectly merged. This process preserves efficiency while enforcing consistent leaf boundaries. Representative example annotations for two classes of the dataset are visualized in \cref{fig:overview_results} and \cref{fig:results_cnw}.

%% file: sec/4_methodology.tex
\section{Methodology}
To identify suitable training configurations and data combinations for leaf segmentation tasks, we first compare the accuracy and efficiency of three model architectures. Based on the most suitable variant, we conduct cross-domain generalization experiments to evaluate the potential of all available datasets for different application scenarios. 

\subsection{Architecture ablation}
\label{sec:architecture_ablation}
We evaluate three architecturally distinct model families on leaf-segmentation data to thoroughly assess their trade‑off between inference speed and segmentation accuracy:

\paragraph{One-stage CNN} YOLO26 \cite{JoCh24} represents a fast convolutional one‑stage baseline. Its architecture follows the principles of CNN‑based real‑time detectors and consists of a backbone for hierarchical feature extraction, a feature‑aggregation neck and a task‑specific prediction head. The model is optimized for real‑time inference on compute‑restricted platforms, making it well suited for mobile or embedded deployment scenarios.

\paragraph{Two-stage CNN} Detectron2 \cite{WuKi19} is used as a high-accuracy baseline, adopting an optimized Mask R‑CNN implementation. The combination of region proposals and ROI‑aligned mask refinement provides strong segmentation performance. To better preserve details of small leaves at low resolutions, we reduce early‑stage backbone strides for low input sizes, preventing overly coarse feature maps.

\paragraph{Vision Transformer} RF-DETR \cite{RoRo25} serves as the third baseline, using a DINOv2 \cite{OqDa23} Transformer backbone and hybrid encoder for fully attention‑based detection and segmentation. Its self‑attention mechanism captures long‑range spatial relationships between leaf structures, leveraging the representational strength of Transformers.

\paragraph{} For the architecture ablation, pre-trained models are fine-tuned on \textit{PhenoBench}, which is the largest and most diverse of our selected datasets, using the respective standard training pipeline of each framework. We use the same split into training, validation and test data with a ratio of 70:15:15 across all experiments. To ensure comparability, models are evaluated using a uniform evaluation protocol based on pycocotools \cite{LiMa14}, thus avoiding framework-specific deviations in metrics calculation.

\subsection{Cross-dataset generalization}
\label{sec:cross_dataset}
To investigate generalizability, the best-performing architecture is evaluated in cross-dataset experiments. For this purpose, we split each of the four selected datasets into training, validation and test sets with a ratio of 70:15:15. We then fine-tune pre-trained weights of the selected architecture on each training set and their combination (\textit{ReLeaf}), resulting in a total of five models to be evaluated. Each one is applied separately to the test set corresponding to its own training data, as well as all samples in each of the remaining datasets, ensuring that training and test data are strictly separated for both in-domain and cross-domain experiments.

\subsection{Hardware setup and training parameters} 
For all experiments, weights pre-trained on the COCO dataset (train2017 and val2017) are fine-tuned for instance segmentation of individual leaves. The experiments investigate different model capacities at four input resolutions (\(192^{2}, 384^{2}, 576^{2}, 768^{2}\) pixels). The resolutions are deliberately chosen as multiples of 32 to account for the maximum downsampling rate of all architectures and thereby ensure that feature maps, anchors, and ROI operations are correctly aligned and no artifacts arise from padding or rounding errors. However, at the lowest resolutions ($192^2$ and $384^2$), a standard stride of 32 would result in feature maps of only 6$\times$6 or 12$\times$12 pixels, which is insufficient for detecting small heart leaves. Therefore, for the Detectron2 baseline, we manually adjust the stride stages (down to 8 and 16) to maintain a higher spatial resolution in the deeper layers of the feature pyramid. Training and evaluation are performed on an NVIDIA RTX 2080 Ti with 11GB VRAM. Due to the limited GPU memory capacity, gradient accumulation is used for the Transformer and two-stage models trained over 50 epochs to increase the effective batch size and improve training stability. All other hyperparameters and augmentation techniques are set according to standard configurations of the respective architectures.

\subsection{Evaluation metrics}
Since precision-agriculture applications typically involve platforms with limited hardware resources, we evaluate model architectures in terms of both their segmentation accuracy and inference speed.

The \textbf{accuracy metric}, as the primary measure of segmentation quality, is calculated using the mean average precision (mAP) based on the COCO standard \cite{LiMa14}, specifically the mAP\textsuperscript{50-95}, which represents the average precision over ten Intersection-over-Union thresholds from 0.50 to 0.95 and is therefore more robust against small segmentation and localization errors, but also more challenging, than the commonly used mAP\textsuperscript{50}.

\textbf{Model latency} is measured per patch as the pure processing time within the respective model API after a warm-up phase of 100 runs. The metric includes the forward pass as well as framework-internal pre- and post-processing steps, including non-maximum suppression.

%% file: sec/5_results.tex
\section{Results}
This section summarizes the central results of our architecture ablation and cross-dataset experiments. More detailed quantitative and qualitative insights are included in the supplementary material.

\begin{figure}
    \centering
    \includegraphics[width=1.0\linewidth]{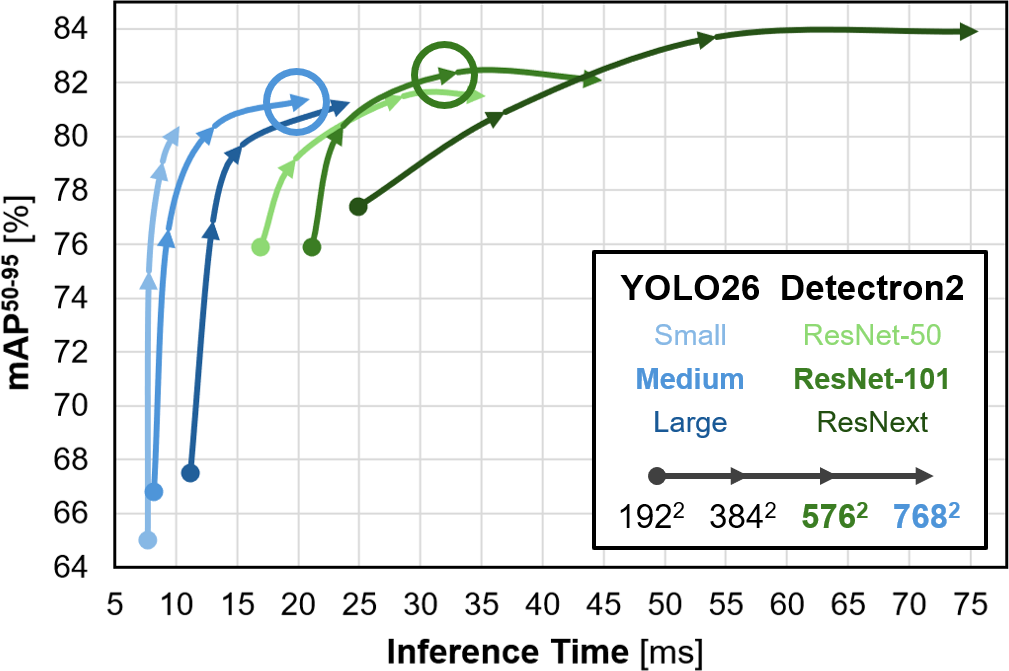}
    \caption{Comparison of leaf-segmentation performance (\% mAP\textsuperscript{50–95}) and inference time (ms) for multiple model-architecture variants and input resolutions on the \textit{PhenoBench} test set. Selected variants providing a reasonable trade-off for real-time deployment are marked by circles.}
    \label{fig:eval_iseg}
\end{figure}

\subsection{Architecture ablation}
To thoroughly compare the model architectures described in \cref{sec:architecture_ablation}, we train and evaluate multiple variants of each and analyze the influence of model complexity and input resolution, as summarized in \cref{fig:eval_iseg}. RF-DETR was omitted from this chart and further analysis, since initial experiments showed accuracy, runtime and training time to be significantly inferior to the other architectures when trained on the selected GPU. This confirms our initial assumption that the hardware requirements of Transformer-based architectures are not readily compatible with systems commonly used in precision-agriculture applications.

\paragraph{Model capacity} Unsurprisingly, increasing model capacity generally leads to improved instance-segmentation performance, but also causes significantly higher inference times. For the YOLO26 models, the accuracy gains are comparatively small, especially when transitioning to larger model variants, while runtime increases almost linearly for all capacities. For the Detectron2 models, however, it is evident that larger variants do not achieve consistent improvements in accuracy despite increasing runtime.

\paragraph{Input resolution} For each architecture and model size, the input resolution was varied in four steps between \(192^{2}\) and \(768^{2}\) pixels. The results show that training on larger images initially leads to performance improvements. However, at high resolutions, starting from \(576^{2}\) pixels, the gains in accuracy become progressively smaller, while runtime increases almost linearly from the outset.

\begin{figure}
    \centering
    \includegraphics[width=1.0\linewidth]{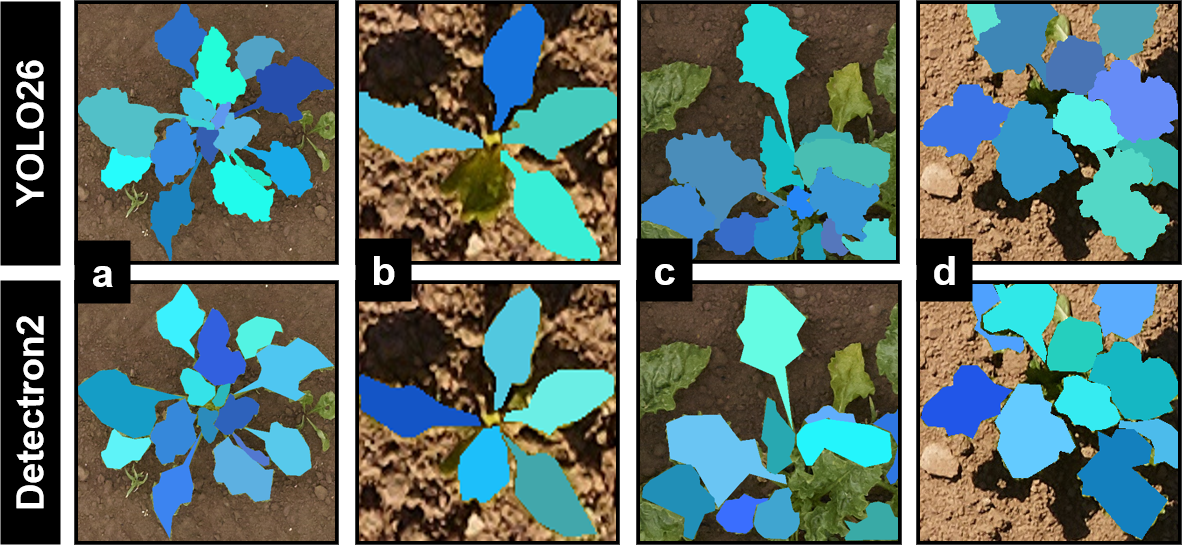}
    \caption{Qualitative leaf-segmentation results on representative test images from \textit{PhenoBench} \cite{WeJa24} using selected model variants of YOLO26 (\textit{Medium} with \(768^{2}\) pixels) and Detectron2 (\textit{ResNet-101} with \(576^{2}\) pixels).}
    \label{fig:qualitative_results_phenobench}
\end{figure}

\paragraph{Accuracy-latency trade-off}
Based on the results of our ablation study, we attempt to define configurations providing a suitable trade-off between segmentation accuracy and inference time for both architectures. In the case of YOLO26, we select the \textit{Medium} model combined with the largest examined resolution, as a higher model capacity does not significantly improve accuracy. Similarly, Detectron2 performs best in combination with \textit{ResNet-101} and an input resolution of \(576^{2}\) pixels. In this case, accuracy actually decreases with higher resolutions and is only surpassed by the \textit{ResNext} variant, which, however, significantly increases runtime. Qualitative results are visualized in \cref{fig:qualitative_results_phenobench} for both of these configurations. Overall, we selected the YOLO26 architecture with the specified configuration for further experiments, since Detectron2 performs only marginally better regarding segmentation accuracy, but drastically increases inference time. 

\subsection{Cross-dataset experiments}
As described in \cref{sec:cross_dataset}, we use the selected YOLO26 configuration to train models on the training splits of all datasets and evaluate performance both on the corresponding test split and on all samples of each remaining dataset, to assess in-domain and cross-domain performance. Additionally, we train a model on a combination of all four training splits incorporating all available information (\textit{ReLeaf}) and compare its results on the corresponding test splits to those from single-dataset models. Furthermore, we evaluate all models on our newly annotated \textit{CropAndWeedAndLeaf} benchmark to analyze generalization to completely different plant species. \cref{tab:cross_evaluation} summarizes the results of these experiments.

\paragraph{In-domain evaluation}
As expected, all models perform reasonably well on their own test sets, as shown by the gray diagonal values. A comparison of these values with each other, however, additionally provides a hint of the complexity and variability provided by each test dataset and an overall measure of the architecture's suitability for different scenarios. The \textit{PhenoBench} model achieves an mAP\textsuperscript{50-95} of 81.4\%, confirming the robustness of the chosen setup for complex, real-world field data. In contrast, the \textit{Komatsuna} model achieves a peak value of 90.7\%, which can be largely attributed to the controlled conditions and the comparatively low visual complexity of the laboratory environment and limited variation in plant appearances. The \textit{LSC} dataset seems to provide higher variation despite its similarly controlled laboratory setting, as its models generalize significantly better to \textit{Komatsuna} data than the other way around. Although both datasets mainly focus on tobacco plants, the addition of a second class and more variable context is apparently beneficial for \textit{LSC}.

\paragraph{Cross-species generalization}
As opposed to the \textit{LSC} and \textit{Komatsuna} datasets, which provide both similar settings and plant species, \textit{PhenoBench} and \textit{GrowliFlower} share the domain of real field imagery but focus on different plant species. A model trained on sugar-beet images from \textit{PhenoBench} achieves an mAP\textsuperscript{50-95} of 49.1\% on the entire \textit{GrowliFlower} dataset, which is focused on cauliflower plants. The reverse transfer results in 38.4\%, again suggesting that a larger image base as provided by \textit{PhenoBench}, is also beneficial for species generalization. This is confirmed when transferring to the 23 plant species in the \textit{CropAndWeedAndLeaf} benchmark, for which the \textit{PhenoBench} model shows superior performance, as well. In general, while all of these values are naturally below the respective in-domain accuracy, they demonstrate the architecture's ability to abstract fundamental morphological leaf traits across species.

\begin{table}
    \caption{Leaf-segmentation performance (\% mAP\textsuperscript{50–95}) of models trained on Lab (\textit{\textbf{KO}matsuna}, \textbf{\textit{LSC}}), Real (\textit{\textbf{P}heno\textbf{B}ench}, \textit{\textbf{G}rowli\textbf{F}lower}) and combined \textit{\textbf{R}e\textbf{L}eaf} datasets. Models are evaluated on their own test splits for reference (\textcolor{gray}{gray}) and cross-dataset on all samples of other sets, including our \textit{\textbf{C}ropAnd\textbf{W}eedAnd\textbf{L}eaf} benchmark. Averages of cross-dataset results are provided for \textbf{Lab} and \textbf{Real} domains.}
    \label{tab:cross_evaluation}
    \centering
    \begin{tabular}{@{}l|rrr|rrrr@{}}
        \toprule
         & LSC & KO & \textbf{Lab} & GF & PB & \textit{CWL} & \textbf{Real} \\
         \midrule
        \textbf{LSC} & \textcolor{gray}{79.2} & 49.1 & \multirow{2}{*}{\textbf{44.7}} & 26.8 & 10.5 & \textit{17.4} & \multirow{2}{*}{\textbf{17.7}}\\
        \textbf{KO} & 40.2 & \textcolor{gray}{90.7} & & 22.7 & 12.3 & \textit{16.3} \\
        \midrule
        \textbf{GF} & 30.2 & 42.5 & \multirow{2}{*}{\textbf{29.8}} & \textcolor{gray}{76.4} & 38.4 & \textit{31.5} & \multirow{2}{*}{\textbf{38.6}}\\
        \textbf{PB} & 40.0 & ~6.6 & & 49.1 & \textcolor{gray}{81.4} & \textit{35.4} \\
        \midrule
        \textbf{RL} & \textcolor{gray}{79.4} & \textcolor{gray}{90.1} & \textbf{\textcolor{gray}{84.8}} & \textcolor{gray}{82.2} & \textcolor{gray}{83.8} & \textit{40.2} & \textbf{\textcolor{gray}{83.3}} \\
        \bottomrule
    \end{tabular}
\end{table}

\paragraph{Cross-domain generalization}
The limits of generalizability become significantly more apparent when transferring between structurally different domains. While the real-world \textit{PhenoBench} model still achieves 40.0\% mAP\textsuperscript{50-95} on the laboratory dataset \textit{LSC}, its performance decreases dramatically to only 6.6\% on \textit{Komatsuna}. This drop can be explained by the controlled laboratory environment, including backgrounds and sensor characteristics, which represent a significant domain barrier for models optimized on noisy field data. \textit{GrowliFlower} achieves better results but still fares well below its in-domain performance. On average, real-world models achieve only 29.8\% when transferred to laboratory settings, as opposed to an average 38.6\% between each other. However, the performance decrease in the opposite direction is even more drastic, with laboratory models achieving an average of only 17.7\% on real-world test data (compared to 44.7\% between each other). This sharp drop underscores the enormous morphological and photometric complexity of field environments, which cannot be adequately modeled by laboratory data alone.

\begin{figure}
    \centering
    \includegraphics[width=1.0\linewidth]{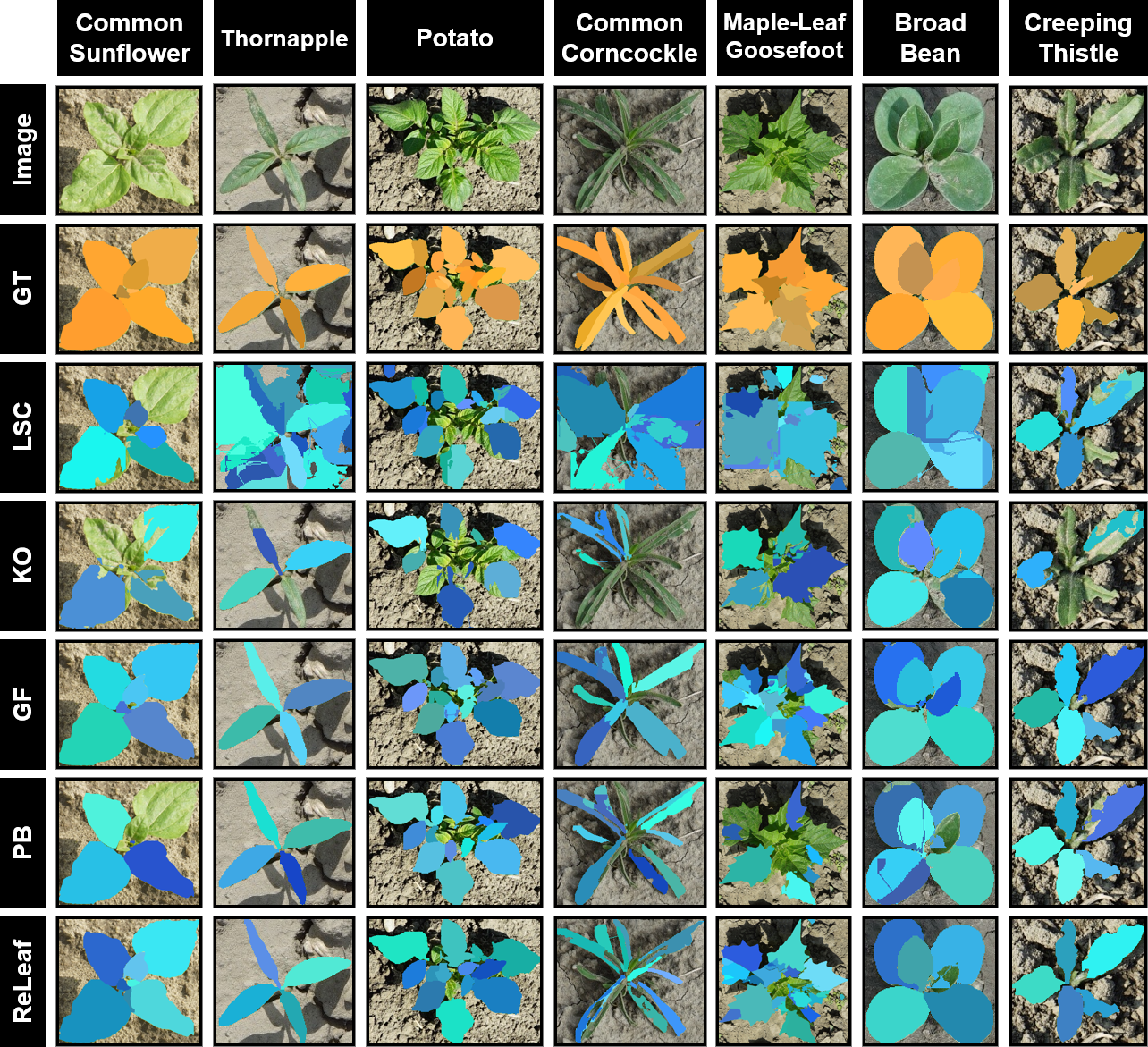}
    \caption{Representative leaf-segmentation results of YOLO26 models (\textit{Medium}, \(768^{2}\) pixels input resolution) trained on \textbf{\textit{LSC}} \cite{ScMi14}, \textit{\textbf{KO}matsuna} \cite{UcSa17}, \textit{\textbf{G}rowli\textbf{F}lower} \cite{KiJu23} and \textit{\textbf{P}heno\textbf{B}ench} \cite{WeJa24} datasets and their combination (\textbf{\textit{ReLeaf}}) on the \textit{CropAndWeedAndLeaf} benchmark, along with corresponding ground truth (\textbf{GT}).}
    \label{fig:results_cnw}
\end{figure}

\paragraph{Multi-domain fusion}
Training a combined \textit{ReLeaf} model on all four datasets proved to be the most effective approach for overcoming this barrier and stabilizing performance across species and domains. This strategy achieves an average overall mAP\textsuperscript{50-95} of 83.9\% across its four corresponding test sets, demonstrating the essential role of integrating diverse morphological traits during training. As the \textit{CropAndWeedAndLeaf} dataset, with its novel classes, was still completely excluded from training, it continues to serve as a benchmark for cross-species generalization. Even in this case, the \textit{ReLeaf} model achieves significantly higher accuracy than those trained on individual datasets, further emphasizing its generalization capability. Overall, the combined approach minimizes performance drops during domain switching and provides increased robustness. \cref{fig:results_cnw} further illustrates this by comparing generalization results of all evaluated models on completely unseen plant species from our \textit{CropAndWeedAndLeaf} benchmark.

\subsection{Discussion}
Segmentation quality and inference speed vary significantly between training resolutions and the different variants of YOLO26 and Detectron2, while RF-DETR proved unsuitable for common real-time precision-agriculture tasks. In particular Detectron2 demonstrated slightly superior precision, presumably related to its capability of segmenting very small leaf instances. The architectural adaptation of its network strides for low input sizes enabled the models to preserve spatial resolution that would otherwise be lost with standard strides of 32. However, Detectron2's two-stage architecture results in significantly longer training and inference times compared to the YOLO26 series, which provides only slightly inferior segmentation quality.

Our cross-domain experiments highlight the performance discrepancy between controlled laboratory environments and unstructured field scenarios specifically for leaf segmentation. While models trained on the \textit{Komatsuna} and \textit{LSC} laboratory datasets achieve peak mAP\textsuperscript{50-95} scores on their own test splits, performance drops drastically when they are transferred to real-world test data. This demonstrates a pronounced domain barrier and confirms the greater complexity of field scenarios, which are characterized by noisy backgrounds, changing light conditions and weed pressure, none of which occur in sterile laboratory environments. A multi-domain \textit{ReLeaf} model combining all training sets successfully overcomes this barrier and establishes a stable performance of 83.9\% across all species and domains in the corresponding test sets and 40.2\% on the unseen species in our \textit{CropAndWeedAndLeaf} benchmark. These results are promising, but may still be insufficient for practical applications depending on the target species. This underlines the need for leaf annotations in large-scale, field-oriented training data such as the \textit{CropAndWeed} dataset.

\subsection{Limitations}
Despite the high overall performance of our pipeline, several limiting factors define the current boundaries of vision-based leaf segmentation. Although patch extraction improves local pixel density, small juvenile leaves remain a challenge, as visible in \cref{fig:results_cnw}. The \textit{PhenoBench} dataset, with an average patch size of 152$\times$141 pixels, reaches mathematical limits with its smallest instances, especially when leaves are partially obscured by soil, other leaves or debris. Even the optimized Detectron2 with reduced strides struggles to detect features in these cases. Annotation quality and data scarcity are also limiting factors, as even human experts find it difficult to define exact instance boundaries for overlapping leaves. Additionally, it should be noted that all experiments were limited by the 11 GB of VRAM available on the NVIDIA RTX 2080 Ti used for training, as we explicitly focus on broad applicability and reproducibility of our results within the research community and in realistic application scenarios. This limitation prevented the evaluation of larger batch sizes and the more memory-intensive Transformer models over longer training periods, which underlines that in practical deployment scenarios, the trade-off between model complexity and the available hardware on an autonomous system remains a critical constraint.

%% file: sec/6_conclusion.tex
\section{Conclusion}
This work addresses the data and research gap in leaf-level instance segmentation, focusing on generalizability under heterogeneous conditions. We conduct a comprehensive investigation of modern deep-learning architectures in this context, showing that YOLO26, particularly its \textit{Small} and \textit{Medium} variants, represents a highly efficient operating point for time-critical applications with over 80\% mAP\textsuperscript{50-95} on the \textit{PhenoBench} dataset, while Detectron2, with a modified network stride, offers superior precision for very small instances at the cost of significantly higher latency.

Based on a YOLO26 architecture providing a suitable trade-off between segmentation accuracy and computational efficiency, we evaluate cross-species and cross-domain generalization of models trained on four publicly available datasets. For this purpose, we created a novel \textit{CropAndWeedAndLeaf} benchmark of 23 plant species by annotating images from the \textit{CropAndWeed} dataset with leaf-level masks. The results demonstrate that abstract morphological leaf traits are successfully learned across species boundaries within the same recording domain, but the shift between controlled laboratory conditions and unstructured field scenarios poses a greater challenge. While laboratory models achieve a peak mAP\textsuperscript{50-95} of 90.7\% in their own domain, they show a drastic performance drop to an average of 17.7\% when transferred to real-world data. Implementing a mixed-domain strategy substantially improves mAP\textsuperscript{50-95} to an average of 83.9\% across all test sets and 40.2\% on our novel benchmark.

As future extensions to this work, we plan to expand our benchmark dataset to a size suitable for training models, since it proved its potential to complement the current data landscape. Regarding our generalization analysis, exploring the potential of unsupervised or semi-supervised domain-adaptation techniques for bridging the lab-to-real gap in this particular domain is a logical next step. Furthermore, evaluating additional deep-learning architectures might be interesting, especially Transformer-based and foundational models combined with extended hardware resources, which were not within the scope of our evaluation.

Overall, this work contributes to identifying and closing key data and domain gaps in leaf segmentation. By combining comprehensive ablation studies with the introduction of a new multi-species benchmark dataset, it offers a substantial step toward more accessible, robust and generalizable perception models.

%% file: sec/X_suppl.tex
\clearpage
\setcounter{page}{1}
\maketitlesupplementary

\appendix
This supplementary document complements the main paper with additional statistics regarding our benchmark dataset and more detailed quantitative and qualitative experimental results to facilitate a more thorough understanding of applied methods and insights.
 
\section{Benchmark dataset}
Our \textit{CropAndWeedAndLeaf} benchmark comprises a representative set of images sampled from 23 classes of the \textit{CropAndWeed} dataset \cite{StTr23}. \cref{tab:cwl_statistics} gives a complete list of the exact species and the \textit{ReLeaf} model's performance on each of them. It shows strong variations between classes related to their appearance and leaf structure, highlighting the overall need for more variable leaf-segmentation data.

\section{Extended results}
The following sections show detailed results of both our ablation and cross-dataset experiments.

\begin{table}
    \caption{Overview of 23 plant species included in \textit{CropAndWeedAndLeaf} with leaf-segmentation annotations for 15 samples each, along with corresponding instance-segmentation performance (\% mAP\textsuperscript{50-95}). Scores are results of the selected model architecture (YOLO26 \textit{Medium}, input size of \(768^{2}\) pixels), trained on the combined \textit{ReLeaf} dataset.}
    \label{tab:cwl_statistics}
    \centering
    \begin{tabular}{@{}llr@{}}
        \toprule
        \textbf{Species name} & \textbf{\textit{Botanical name}} & \textbf{Score} \\
        \midrule
        Black-bindweed & \textit{Fallopia convolvulus} & 41.6 \\
        Broad bean & \textit{Vicia faba} & 49.7 \\
        Common bean & \textit{Phaseolus vulgaris} & 66.5 \\
        Common corncockle & \textit{Agrostemma githago} & 20.2 \\
        Common sunflower & \textit{Helianthus annuus} & 73.0 \\
        Copse bindweed & \textit{Fallopia dumetorum} & 43.9 \\
        Cornflower & \textit{Centaurea cyanus} & 30.0 \\
        Creeping thistle & \textit{Cirisium arvense} & 39.9 \\
        Field sowthistle & \textit{Sonchus arvensis} & 34.7 \\
        Maize & \textit{Zea mays} & 41.2 \\
        Maple-leaf goosefoot & \textit{Chenopodium hybridum} & 31.7 \\
        Pea & \textit{Pisum sativum} & 11.2 \\
        Poppy & \textit{Papaver} & 54.5 \\
        Potato & \textit{Solanum tuberosum} & 32.5 \\
        Red-root amaranth & \textit{Amaranthus retroflexus} & 57.9 \\
        Redshank & \textit{Persicaria maculosa} & 26.7 \\
        Ribwort plantain & \textit{Plantago lanceolata} & 34.8 \\
        Soybean & \textit{Glycine max} & 27.2 \\
        Squash & \textit{Cucurbita} & 60.9 \\
        Sugar beet & \textit{Beta vulgaris s. vulgaris} & 66.0 \\
        Thornapple & \textit{Datura stramonium} & 57.6 \\
        White goosefoot & \textit{Chenopodium album} & 39.4 \\
        Zucchini & \textit{Cucurbita pepo var. gir.} & 68.2 \\
        \bottomrule
    \end{tabular}
\end{table}

\subsection{Architecture ablation}
As stated in the main paper, RF-DETR was omitted from the final ablation after initial experiments showed drastically inferior performance compared to other architectures under the applied hardware constraints. \cref{tab:ablation_architectures} provides an overview of these initial results. While inference times of RF-DETR are comparable to those of YOLO26 and even slightly lower than for Detectron2, mAP values are significantly inferior to both. At the same time, training times are higher by a factor of 3.5 compared to Detectron2 at the same image resolution and 6.3 compared to YOLO26 models trained at even higher resolutions, which are not available for RF-DETR due to its substantial VRAM requirements. Mitigating these discrepancies and making transformer-based architectures usable for our hardware and data setup could still be achieved by using specialized architecture variants (e.g., RT-DETR \cite{LyZh24}, Deformable DETR \cite{ZhWe21}) or dedicated training strategies, such as memory-efficient attention mechanisms or staged training starting from plant-level instance-segmentation datasets. These options may be worth exploring in future extensions of this work, but were not within the scope of our analysis.

\begin{table}
    \caption{Initial ablation results for RF-DETR \textit{SegPreview} with multiple input resolutions, including mask accuracy as well as training and inference times (training at an input resolution of $768^2$ pixels was not feasible due to high VRAM requirements). Results of selected YOLO26 and Detectron2 variants (\textit{Medium} and \textit{ResNet-101}, respectively) are provided for reference. All models are trained for 20 epochs using the \textit{PhenoBench} training/validation splits and evaluated on the corresponding test set.}
    \label{tab:ablation_architectures}
    \begin{tabular}{@{}lc|ccc@{}}
        \toprule
         & \textbf{Size} & \textbf{Accuracy} & \textbf{Training} & \textbf{Inference} \\
         & [px] & [\% mAP] & [h/epoch] & [ms/image] \\
         \midrule
        \multirow{3}{*}{\textbf{RF-DETR}} & $192^2$ & 55.9 & 0.85 & 18.0 \\
         & $384^2$ & 59.2 & 0.94 & 19.7 \\
         & $576^2$ & 59.9 & 1.13 & 23.4 \\
         \midrule
        \textit{\textbf{Detectron2}} & \textit{576\textsuperscript{2}} & \textit{80.0} & \textit{0.32} & \textit{33.1} \\
        \textit{\textbf{YOLO26}} & \textit{768\textsuperscript{2}} & \textit{78.0} & \textit{0.18} & \textit{21.0} \\
        \bottomrule
    \end{tabular}
\end{table}

\begin{figure}
    \centering
    \includegraphics[width=1.0\linewidth]{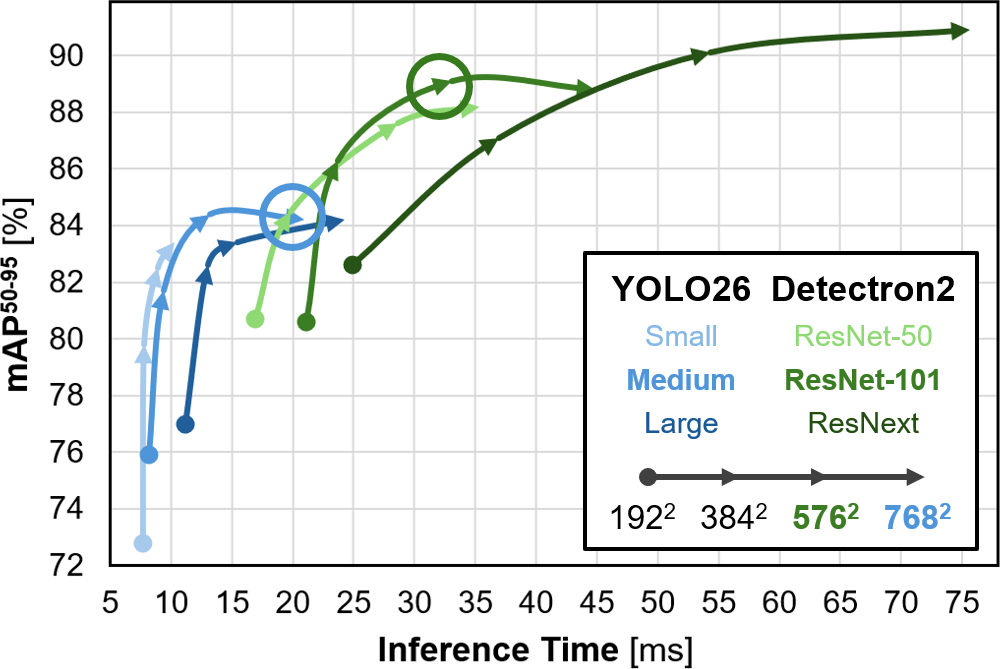}
    \caption{Comparison of bounding-box accuracy (\% mAP\textsuperscript{50–95}) and inference time (ms) of leaf-segmentation for multiple model-architecture variants and input resolutions on the \textit{PhenoBench} test set. Selected variants providing a reasonable trade-off for real-time deployment are marked by circles.}
    \label{fig:eval_det}
\end{figure}

In addition to the accuracy of leaf-segmentation masks provided in \cref{fig:eval_iseg} of the main paper, we present the corresponding bounding-box accuracy in \cref{fig:eval_det} to allow a differentiation between the models' capabilities of correctly localizing individual leaves and accurately delineating their contours. As visible in the results, the discrepancy is significantly larger for Detectron2 \cite{WuKi19} compared to YOLO26 \cite{JoCh24}, indicating that it is a superior choice for applications requiring pure detection in the form of bounding boxes without fine-grained object masks. For the purposes investigated in this work, however, YOLO26 remains a more suitable choice due to its lower processing time. \cref{tab:ablation_results} shows the exact values forming the basis of both charts.

\begin{table}
    \caption{Comparison of bounding-box (\textbf{Box}) and segmentation-mask (\textbf{Mask}) accuracies (\% mAP\textsuperscript{50–95}) and inference times (\textbf{t}) (ms) of leaf-segmentation for multiple model-architecture variants and input resolutions on the \textit{PhenoBench} test set. Selected variants providing a reasonable trade-off for real-time deployment are highlighted by bold text.}
    \label{tab:ablation_results}
    \centering
    \begin{tabular}{@{}ccc|ccc@{}}
        \toprule
        \multicolumn{1}{l}{} & \multicolumn{1}{l}{} & \textbf{Size} & \textbf{Box} & \textbf{Mask} & \textbf{t} \\
        \midrule
        \multirow{12}{*}{\rotatebox{90}{\textbf{YOLO26}}} & \multirow{4}{*}{\textbf{Small}} & $192^2$ & 72.8 & 65.0 & ~~7.7 \\
         &  & $384^2$ & 79.8 & 75.0 & ~~7.8 \\
         &  & $576^2$ & 82.5 & 79.1 & ~~8.9 \\
         &  & $768^2$ & 83.4 & 80.4 & 10.3 \\
         \cmidrule(l){2-6}
         & \multirow{4}{*}{\textbf{Medium}} & $192^2$ & 75.9 & 66.8 & ~~8.2 \\
         &  & $384^2$ & 81.7 & 76.6 & ~~9.4 \\
         &  & $576^2$ & 84.4 & 80.4 & 13.2 \\
         &  & \textbf{768\textsuperscript{2}} & \textbf{84.2} & \textbf{81.4} & \textbf{21.0} \\
         \cmidrule(l){2-6}
         & \multirow{4}{*}{\textbf{Large}} & $192^2$ & 77.0 & 67.5 & 11.2 \\
         &  & $384^2$ & 82.6 & 76.9 & 13.1 \\
         &  & $576^2$ & 83.4 & 79.7 & 15.5 \\
         &  & $768^2$ & 84.2 & 81.3 & 24.3 \\
         \midrule
        \multirow{12}{*}{\rotatebox{90}{\textbf{Detectron2}}} & \multirow{4}{*}{\textbf{ResNet-50}} & $192^2$ & 80.7 & 75.9 & 16.9 \\
         &  & $384^2$ & 84.5 & 79.2 & 19.9 \\
         &  & $576^2$ & 87.6 & 81.5 & 28.6 \\
         &  & $768^2$ & 88.2 & 81.5 & 35.3 \\
         \cmidrule(l){2-6}
         & \multirow{4}{*}{\textbf{ResNet-101}} & $192^2$ & 80.6 & 75.9 & 21.1 \\
         &  & $384^2$ & 86.3 & 80.4 & 23.7 \\
         &  & \textbf{576\textsuperscript{2}} & \textbf{89.1} & \textbf{82.4} & \textbf{33.1} \\
         &  & $768^2$ & 88.8 & 82.1 & 44.9 \\
         \cmidrule(l){2-6}
         & \multirow{4}{*}{\textbf{ResNext}} & $192^2$ & 82.6 & 77.4 & 24.9 \\
         &  & $384^2$ & 87.1 & 80.9 & 37.0 \\
         &  & $576^2$ & 90.1 & 83.7 & 54.3 \\
         &  & $768^2$ & 90.9 & 83.9 & 75.7 \\
         \bottomrule
    \end{tabular}
\end{table}

\begin{table}
    \caption{Comparison of bounding-box (\textbf{Box}) and segmentation-mask (\textbf{Mask}) accuracies (mAP\textsuperscript{50–95}) for varying objects sizes (\textbf{S}mall $<$ $32^2$ pixels $<$ \textbf{M}edium $<$ $96^2$ pixels $<$ \textbf{L}arge) using the selected configurations of YOLO26 (\textit{Medium} with an input resolution of \(768^{2}\) pixels) and Detectron2 (\textit{ResNet-101 }with an input resolution of \(576^{2}\) pixels).}
    \label{tab:ablation_size}
    \centering
    \begin{tabular}{@{}l|lll|lll@{}}
        \toprule
         & \multicolumn{3}{|c}{\textbf{Box}} & \multicolumn{3}{|c}{\textbf{Mask}} \\
         & \multicolumn{1}{|c}{\textbf{S}} & \multicolumn{1}{c}{\textbf{M}} & \multicolumn{1}{c}{\textbf{L}} & \multicolumn{1}{|c}{\textbf{S}} & \multicolumn{1}{c}{\textbf{M}} & \multicolumn{1}{c}{\textbf{L}} \\
         \midrule
        \textbf{YOLO26} & 69.9 & 82.2 & 91.3 & 69.5 & 79.3 & 89.1 \\
        \textbf{Detectron2} & 87.4 & 88.0 & 91.7 & 80.9 & 80.9 & 86.3 \\
        \bottomrule
    \end{tabular}
\end{table}

\cref{tab:ablation_size} shows the performance of our selected configurations of YOLO26 and Detectron2 on different leaf sizes. It confirms our prior assumption that two-stage detectors like Detectron2 excel at retrieving and accurately delineating small objects at the cost of higher parameter counts and therefore longer inference times than single-stage variants.

\newpage
\subsection{Cross-dataset experiments}
\begin{figure}
    \centering
    \includegraphics[width=1.0\linewidth]{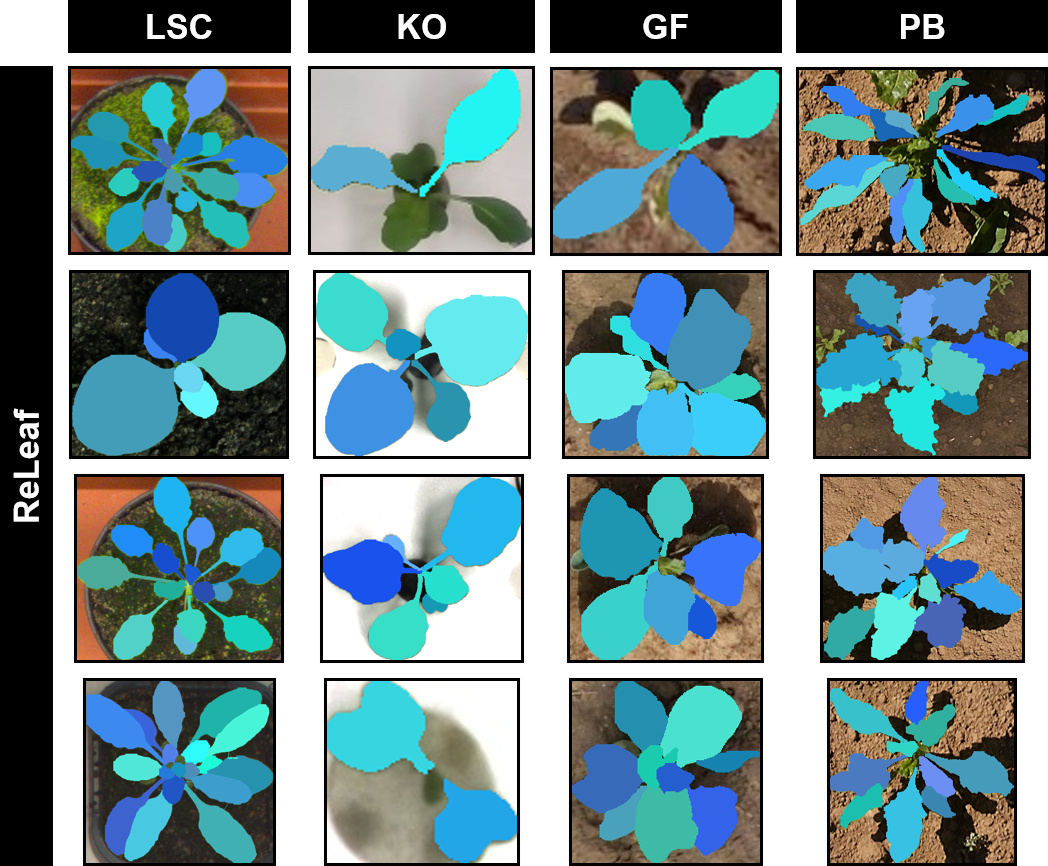}
    \caption{Representative leaf-segmentation results of selected YOLO26 configuration (\textit{Medium} with an input resolution of \(768^{2}\) pixels) trained on a combination of \textit{LSC} \cite{ScMi14}, \textit{Komatsuna} (KO) \cite{UcSa17}, \textit{GrowliFlower} (GF) \cite{KiJu23} and \textit{PhenoBench} (PB) \cite{WeJa24} datasets and evaluated on the respective test sets.}
    \label{fig:results_all}
\end{figure}

\cref{fig:results_all} visualizes our multi-domain \textit{ReLeaf} model's results on a selection of representative test images from \textit{LSC} \cite{ScMi14}, \textit{Komatsuna} \cite{UcSa17}, \textit{GrowliFlower} \cite{KiJu23} and \textit{PhenoBench} \cite{WeJa24}, confirming its stable performance across both species and environmental conditions.

\begin{figure*}
    \centering
    \includegraphics[width=0.98\linewidth]{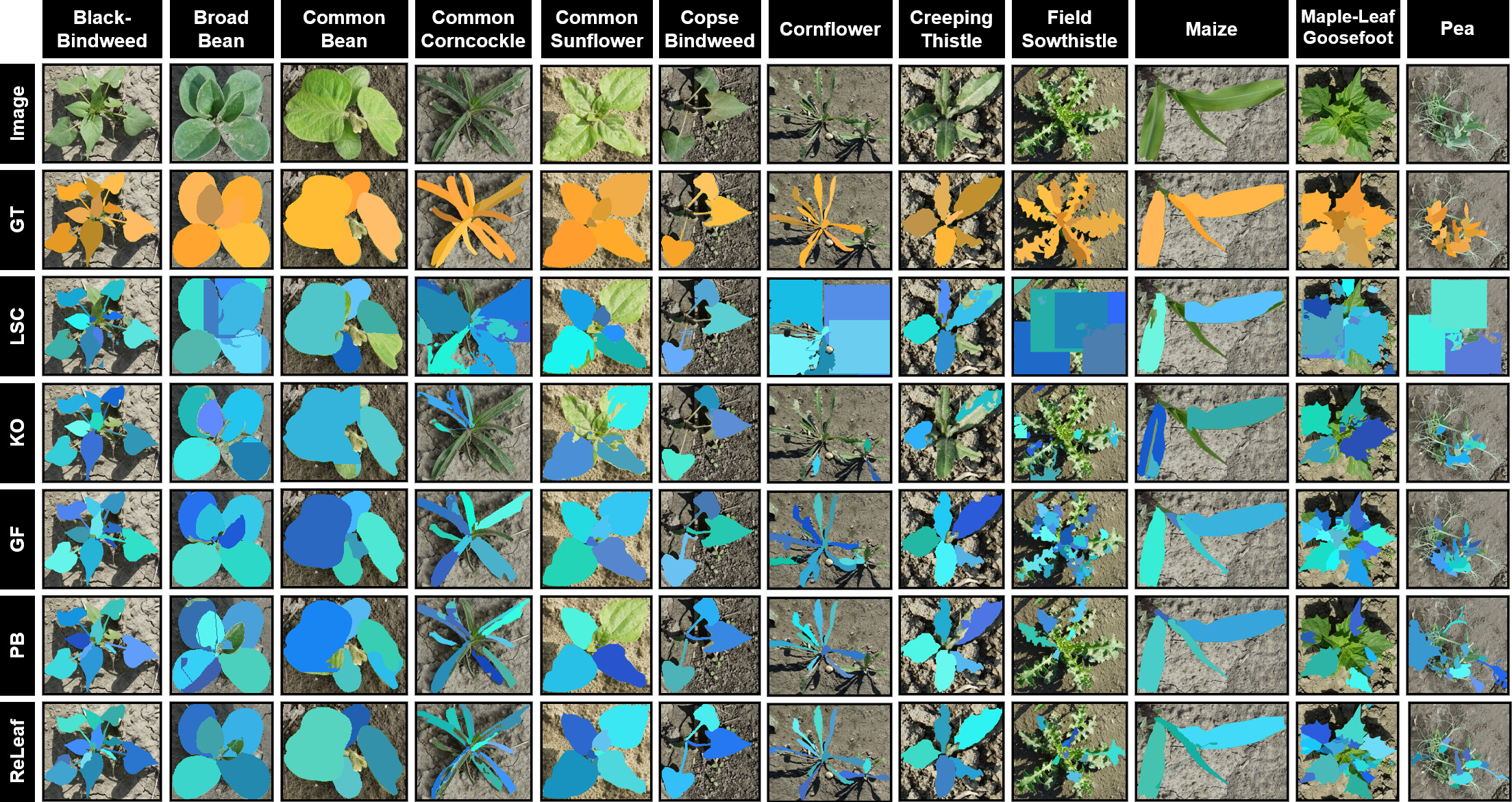}
    \caption{Representative leaf-segmentation results of YOLO26 models (\textit{Medium}, \(768^{2}\) pixels input resolution) trained on \textbf{\textit{LSC}} \cite{ScMi14}, \textit{\textbf{KO}matsuna} \cite{UcSa17}, \textit{\textbf{G}rowli\textbf{F}lower} \cite{KiJu23} and \textit{\textbf{P}heno\textbf{B}ench} \cite{WeJa24} datasets and their combination (\textbf{\textit{ReLeaf}}), evaluated on all 23 species of the \textit{CropAndWeedAndLeaf} benchmark, along with corresponding ground truth (\textbf{GT}) (Part 1).}
    \label{fig:results_cnw_part1}
\end{figure*}

\begin{figure*}
    \centering
    \includegraphics[width=0.89\linewidth]{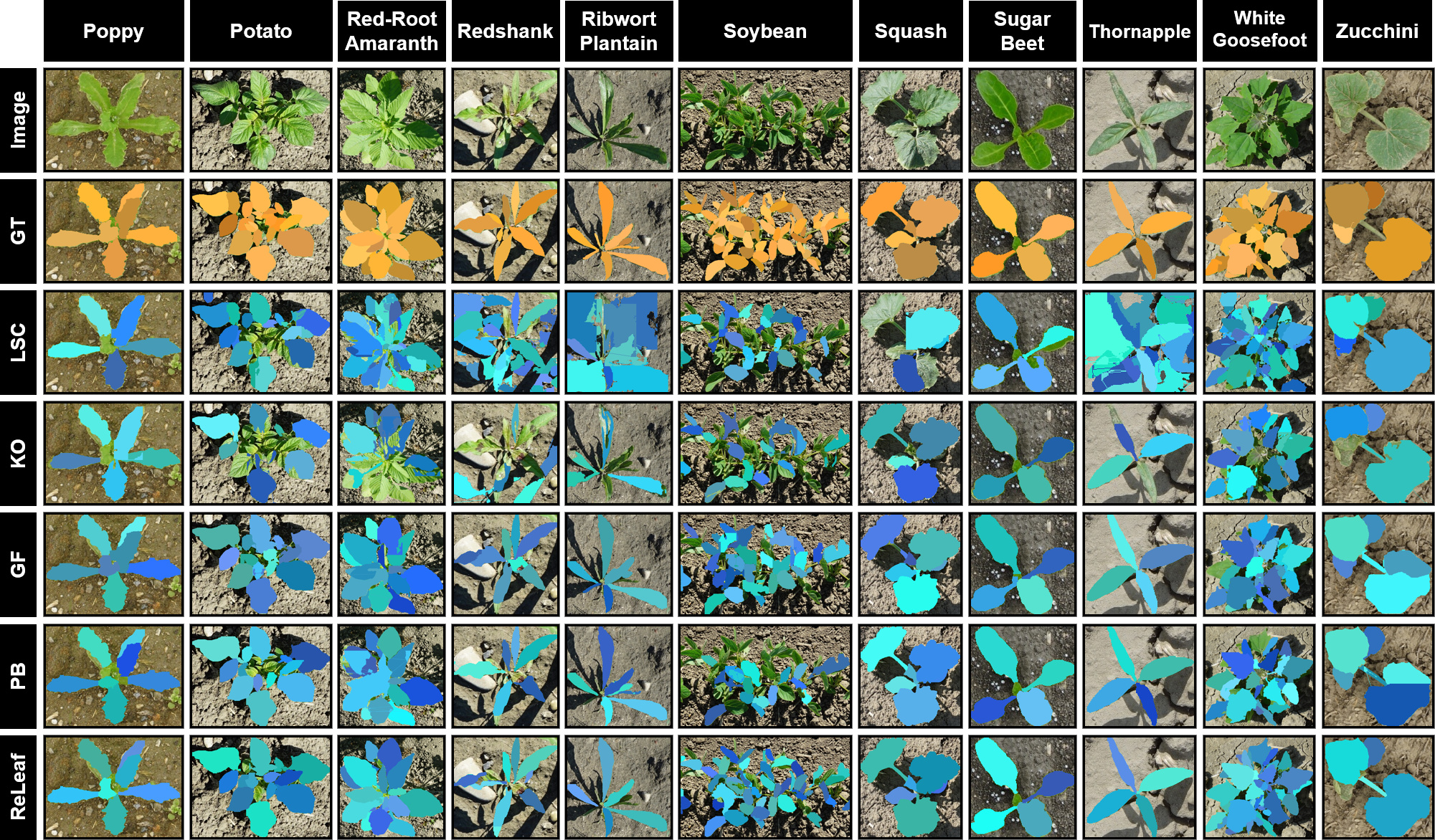}
    \caption{Representative leaf-segmentation results of YOLO26 models (\textit{Medium}, \(768^{2}\) pixels input resolution) trained on \textbf{\textit{LSC}} \cite{ScMi14}, \textit{\textbf{KO}matsuna} \cite{UcSa17}, \textit{\textbf{G}rowli\textbf{F}lower} \cite{KiJu23} and \textit{\textbf{P}heno\textbf{B}ench} \cite{WeJa24} datasets and their combination (\textbf{\textit{ReLeaf}}), evaluated on all 23 species of the \textit{CropAndWeedAndLeaf} benchmark, along with corresponding ground truth (\textbf{GT}) (Part 2).}
    \label{fig:results_cnw_part2}
\end{figure*}

\cref{fig:results_cnw_part1,fig:results_cnw_part2} show the results of models trained on each dataset and evaluated on one representative image for each of the 23 plant species in the \textit{CropAndWeedAndLeaf} benchmark. Models trained on laboratory data (\textit{LSC}, \textit{Komatsuna}) struggle most to generalize to novel plant species in real-world settings. While the combined \textit{ReLeaf} model occasionally misses small inner leaves, it significantly improves adaptation to the strongly varying visual appearances and leaf configurations of different crop and weed species.

\newpage

